\documentclass[letterpaper]{article} 
\usepackage{aaai2026}  
\usepackage{times}  
\usepackage{helvet}  
\usepackage{courier}  
\usepackage[hyphens]{url}  
\usepackage{graphicx} 
\usepackage{natbib}  
\usepackage{caption} 
\usepackage{algorithm}
\usepackage{algorithmic}

\usepackage{graphicx}

\usepackage{amsmath,amssymb,amsfonts}
\usepackage{algorithmic}
\usepackage{graphicx}
\usepackage{textcomp}
\usepackage{xcolor}
\usepackage{amsmath}
\usepackage{dsfont}
\usepackage{multirow}
\usepackage{multicol}
\usepackage{hhline}
\usepackage{subcaption}

\usepackage{newfloat}
\usepackage{listings}
\DeclareCaptionStyle{ruled}{labelfont=normalfont,labelsep=colon,strut=off} 
\floatstyle{ruled}
\newfloat{listing}{tb}{lst}{}
\floatname{listing}{Listing}
\nocopyright

\title{Vehicle Routing with Finite Time Horizon using Deep Reinforcement Learning with Improved Network Embedding}
\author{
    Ayan Maity\textsuperscript{\rm 1}, Sudeshna Sarkar\textsuperscript{\rm 2}
}
\affiliations{
    \textsuperscript{\rm 1}Department of Computer Science and Engineering, IIT Kharagpur\\
    Email: ayanmaity201@kgpian.iitkgp.ac.in\\
    \textsuperscript{\rm 2}Department of Computer Science and Engineering, IIT Kharagpur\\
    Email: sudeshna@cse.iitkgp.ac.in

}

\begin{document}
\maketitle

\section{Abstract}

In this paper, we study the vehicle routing problem with a finite time horizon. In this routing problem, the objective is to maximize the number of customer requests served within a finite time horizon. We present a novel routing network embedding module which creates local node embedding vectors and a context-aware global graph representation. The proposed Markov decision process for the vehicle routing problem incorporates the node features, the network adjacency matrix and the edge features as components of the state space. We incorporate the remaining finite time horizon into the network embedding module to provide a proper routing context to the embedding module. We integrate our embedding module with a policy gradient-based deep Reinforcement Learning framework to solve the vehicle routing problem with finite time horizon. We trained and validated our proposed routing method on real-world routing networks, as well as synthetically generated Euclidean networks. Our experimental results show that our method achieves a higher customer service rate than the existing routing methods. Additionally, the solution time of our method is significantly lower than that of the existing methods.

\section{Introduction}
Vehicle Routing Problems (VRP) are a widely recognized class of combinatorial optimization problems that have been extensively studied \cite{Baty_Jungel_Klein_Parmentier_Schiffer_2024}. Vehicle routing has applications in transportation, logistics, delivery systems and urban planning. 

Different variants of the vehicle routing problem (VRP) have been studied in the past, such as Capacitated VRP \cite{toth2014vehicle}, VRP with finite horizon \cite{Zhang_Luo_Florio_Van_Woensel_2023}, electric vehicle routing problem \cite{tang2023energy}, VRP with time windows \cite{schneider2014electric}, etc. 

In this study, we consider the Vehicle Routing Problem with a Finite Time Horizon (VRP-FTH) \cite{Zhang_Luo_Florio_Van_Woensel_2023}. In the VRP-FTH, a single vehicle departs from a depot, serves a subset of customers, and returns to the depot within a finite service time. Since not all customers can be visited within this time horizon, the objective is to find a trip route to maximize the number of customers served. Customer requests may be deterministic (known beforehand) or stochastic (received during the trip).

Until recently, most of the existing methods to solve the VRP-FTH used traditional optimization methods. The traditional optimization-based routing method include genetic algorithm \cite{wang2008using}, integer program models \cite{klapp2018dynamic}, tabu search-based metaheuristic \cite{ferrucci2015general}, approximate dynamic programming algorithm \cite{kullman2021electric}, multiple knapsack approximation \cite{Zhang_Luo_Florio_Van_Woensel_2023} etc. However, exact methods are computationally expensive, while heuristic methods, though faster, typically yield suboptimal solutions \cite{tang2023energy}.

Recent studies have explored reinforcement learning (RL)-based routing models \cite{kool2018attention, lin2021deep, gama2021reinforcement} for the VRP-FTH. \citet{lin2021deep} proposed a structure2vec-based framework, while \citet{kool2018attention} introduced a multi-head attention model to solve different variants of the VRP. The existing RL-based methods only create local node embeddings and they do not generate a global graph embedding based on the current routing context. Additionally, most of the existing methods rely solely on Euclidean networks and vertex coordinates, which fail to capture real-world graph structures where paths are not necessarily straight lines.

In this paper, we propose a graph attention networks-based routing network embedding module which takes the graph adjacency matrix and the edge features into account to produce the node embedding vectors and the global graph embedding vector. We integrate the routing network embedding module with a policy network-based routing agent to solve the VRP-FTH instances.

The major contributions of this paper are as follows:
\begin{enumerate}
    \item We propose a novel routing network embedding module which generates local node encoding vectors along with a global graph representation vector to facilitate better understanding of the current routing context. Our proposed network embedding module consists of Graph Attention Networks with edge features and Cross-Attention. 
    \item We incorporate the remaining time horizon into the graph embedding module to improve the quality of the global graph representation. 
\end{enumerate}

\begin{figure}
    \centering
    \includegraphics[width=0.7\linewidth]{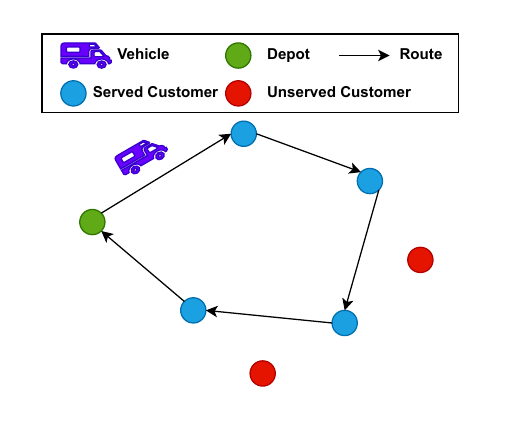}
    \caption{Vehicle Routing with Finite Time Horizon}
    \label{vrp_fth_fig}
\end{figure}

\section{Problem Definition}

The vehicle routing problem with finite time horizon (VRP-FTH) \cite{Zhang_Luo_Florio_Van_Woensel_2023} is defined on a complete and directed routing network $G=(V,E)$, where $V$ is the set of nodes and $E$ is the set of edges. The set of nodes $V$ contains a depot ($v_d$) and a set of customer nodes ($\mathcal{C}$). Each edge $(i,j)$ has a travel time $t_{i,j}$ associated with it. A single vehicle starts at the depot, visits a number of customers and returns to the depot within a finite time horizon $U$. A set of deterministic customers $\mathcal{C}_d$ is known at the start of the trip. New customer requests are received during the trip till $K$ timesteps. The objective is to visit and serve as many customers as possible and return to the depot within the finite horizon $U$. If $i_t$ represents the node visited at timestep $t$, $T$ is the total number of timesteps and $\mathcal{R}=(i_1,i_2,...,i_T)$ represents the route followed by the vehicle, the objective function is mathematically expressed as:
\begin{equation}
\max_{\mathcal{R}{=(i_1,...,i_T)}} \sum_{t=1}^T \mathds{1} [i_t \in \mathcal{C}] \ \text{subject to} \ \sum_{j=1}^{T-1} t_{i_{j},i_{j+1}} \leq U 
\end{equation}
where the function $\mathds{1}[.]$ is an indicator function. Figure \ref{vrp_fth_fig} shows an example of the VRP-FTH.

\section{Related Work}

The VRP-FTH has been extensively studied in previous works, often under different names. In particular, the VRP-FTH with deterministic customers has been referred to as the Orienteering Problem \cite{vansteenwegen2019orienteering}, the Selective Travelling Salesman Problem \cite{derya2020selective} etc. On the other hand, the VRP-FTH with stochastic customers has been referred to as the Dynamic Vehicle Routing Problem with Stochastic Requests \cite{Zhang_Luo_Florio_Van_Woensel_2023}, the Dynamic Customer Acceptance in Delivery Routing \cite{ulmer2020meso}, etc.

We can broadly categorize different vehicle routing methods into two categories: \textit{Traditional Optimization-based Methods} and \textit{Reinforcement Learning-based methods}. 

\subsection{Traditional Optimization-based Routing Methods}

Until recently, most of the vehicle routing methods were based on traditional optimization techniques.

\subsubsection{Traditional Methods for VRP with Finite Horizon.}

\citet{kara2016new} studied the vehicle routing problem with finite horizon and proposed an integer linear programming program model to solve the routing problem. However, they only considered small-scale routing instances. \citet{ferrucci2015general} proposed a tabu search-based metaheuristic algorithm for dynamic vehicle routing with time horizon and time windows. However, this tabu search-based method is not suitable for large-scale routing instances.

\citet{ulmer2020meso} addressed vehicle routing for customer acceptance using a meso-parametric value function approximation method. \citet{ulmer2020modeling} formulated the stochastic dynamic vehicle routing using a route-based Markov decision process (MDP). As a result of the route-based MDP, the state space size increases exponentially with the inclusion of the route plan. 

\citet{urrutia2021variable} presented a variable neighbourhood search algorithm to solve the VRP with finite horizon. On large routing instances, the variable neighbourhood search algorithm is highly prone to get stuck in local optimal values.

\citet{Zhang_Luo_Florio_Van_Woensel_2023} addressed dynamic vehicle routing with stochastic requests using a route-based MDP and a multiple knapsack-based value function approximation. However, the route-based MDP leads to a large state space and high solution time.

\citet{xu2024adapted} proposed a genetic algorithm-based routing method to solve the orienteering problem. However, the solution time of the genetic algorithm is relatively high as compared to the reinforcement learning-based methods.

\subsubsection{Traditional Methods for Other VRP Variants.}
\citet{chen2020electric} considered the vehicle routing problem with time windows and proposed an adaptive large neighbourhood search algorithm to solve the problem. \citet{louati2021mixed} proposed mixed integer programming models to solve vehicle routing problems with pickup and delivery. However, the solution time is impractical for real-time use.

\citet{jia2021bilevel} formulated the capacitated electric vehicle routing problem as a bilevel optimization problem and presented an improved ant colony optimization method to solve the problem. \citet{maroof_ga_2024} presented a hybrid genetic algorithm for vehicle routing problems with time windows. For large routing networks, the genetic algorithm-based methods generally produce sub-optimal results.

\subsection{Reinforcement Learning-based Routing Methods}
In recent times, Reinforcement Learning (RL) has emerged as a promising approach to solve sequential decision making problems. 

\subsubsection{RL Methods for VRP with Finite Horizon.}
\citet{kool2018attention} proposed a multi-head attention model with the REINFORCE method for vehicle routing. However, relying solely on node coordinates limits its ability to represent real road networks, where paths are not necessarily straight lines.

\citet{lin2021deep} proposed a deep reinforcement learning-based method to solve the electric vehicle routing problem with time windows. Their proposed structure2vec (S2V) node embedding module only considers the node coordinates and binary adjacency matrix, which reduces the capacity of the model to represent real road network. 

\citet{gama2021reinforcement} presented a pointer network-based policy gradient method to solve the vehicle routing with time windows and finite time horizon. However, similar to \citet{lin2021deep}, they only considered Euclidean networks and the node coordinates are used as input features.

\citet{hildebrandt2023opportunities} proposed a policy gradient RL framework for the stochastic dynamic vehicle routing problem. However, the absence of network encoding limits its ability to effectively capture the routing network structure.

\subsubsection{RL Methods for Other VRP Variants.}

\begin{figure*}[htbp]
    \centering
    \includegraphics[width=1.0\linewidth]{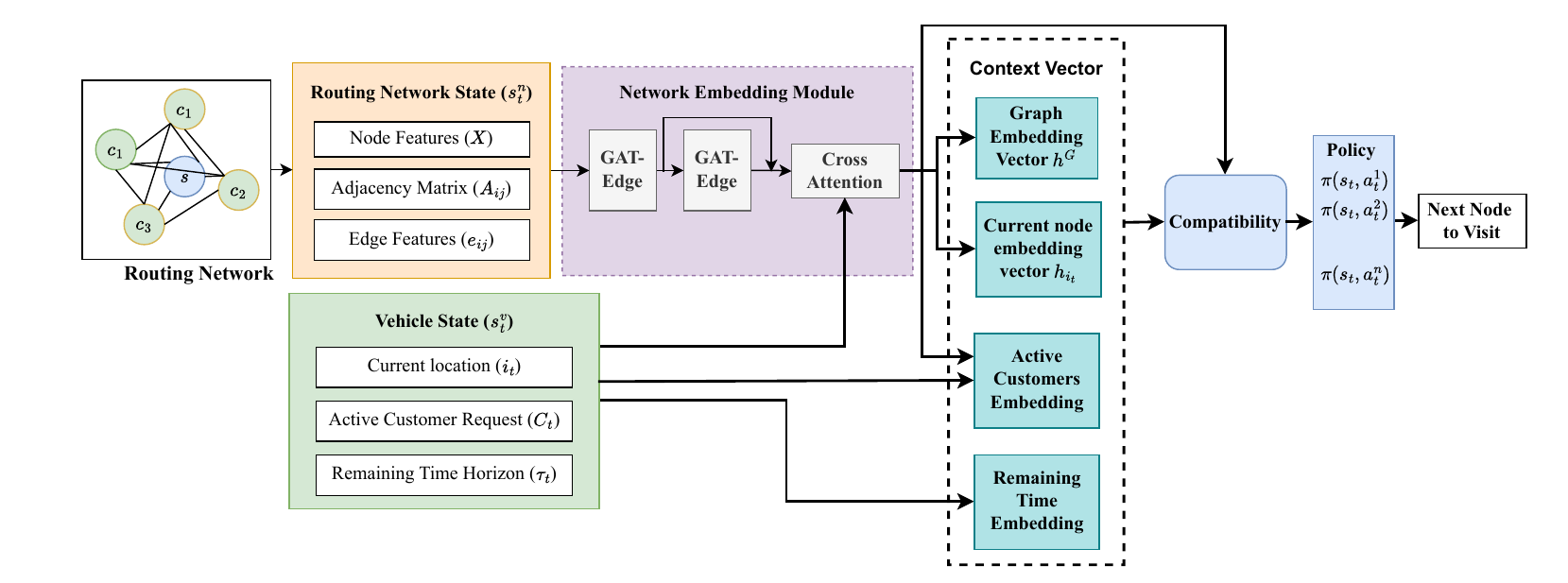}
    \caption{Policy Network with GAT-based routing network embedding module (PG-GAT-Edge)}
    \label{PG_GAT_Edge_fig}
\end{figure*}

\citet{basso2022dynamic} considered the dynamic stochastic vehicle routing problem and presented a tabular safe reinforcement learning method to solve the problem. However, this method is only applicable on small routing instances due to its memory and time requirement.

\citet{tang2023energy} developed a transformer-based policy gradient approach for energy-efficient EV routing, but it is restricted to Euclidean networks and ignores real-world edge features.

\citet{mozhdehi2024edge} proposed an edge-enhanced transformer-based graph embedding model for the electric vehicle routing problem with time windows. \citet{maity2025electric} introduced a Deep Q-Network–based electric vehicle routing approach employing a graph convolutional network for embedding.

Traditional heuristic methods often yield suboptimal solutions, while exact methods are computationally expensive. Most RL-based routing approaches are limited to Euclidean networks, ignoring network structure. Moreover, the existing RL-based routing methods do not produce useful context-aware global graph representation vectors to help the agent understand the global routing context.

\section{Methodology}
\label{method_sec}

This section presents our proposed graph attention networks-based routing network embedding module and the policy network-based routing framework. Figure \ref{PG_GAT_Edge_fig} shows our proposed routing framework. 

The vehicle routing problem with finite time horizon (VRP-FTH) is formulated as a Markov decision process (MDP). In our MDP for the VRP-FTH, the state space incorporates the routing network state which contains node features, sparse adjacency matrix and edge features.

We propose a novel routing network embedding module which creates the node embedding vectors as well as a context-aware global graph embedding vector.

Finally, we present a policy network consisting of the network embedding module, the context vector and the compatibility layer to solve the VRP-FTH.

\subsection{Markov Decision Process}

\label{mdp_sec}

The VRP-FTH is formulated as a Markov Decision Process (MDP) with defined states, actions, and rewards. Unlike existing deep RL methods that use only node coordinates \cite{tang2023energy, kool2018attention}, our formulation incorporates node features, a sparse adjacency matrix, and edge features to better capture the network structure.

\begin{enumerate}
    \item \textit{State}: The state of the agent ($s_t$) contains the routing network state ($s^n_t$) and the  vehicle state ($s^v_t$): $s_t = \{s^n_t,s^v_t\}$. The routing network state ($s^n_t$) consists of three components:
    \begin{enumerate}
        \item Node features ($X$): The node feature of a node $v$ contains the travel time costs to reach the other nodes from $v$ and it also contains whether $v$ is the depot or a customer node.
        \item Adjacency matrix ($A$): As the routing network in our problem is a complete and directed graph, we create a binary adjacency matrix using k-nearest neighbours (KNN) of each node. 
        \item Edge features ($e$): In our problem, the edge feature $e_{ij}$ for the edge $(i,j)$ is computed as: $e_{ij} = 1 - t'_{ij}$, where $t'_{ij}$ is the min-max normalized travel time for the edge $(i,j)$.
    \end{enumerate}
    The routing network state is: $s^n_t = (X, A, e)$.
    
    The current vehicle state is composed of the current location ($i_t$), the set of active customer requests ($C_t$) and the remaining time horizon ($\tau_t$): $s^v_t = (i_t, C_t, \tau_t)$.
    
    \item \textit{Action}:  The action $a_t$ at a state $s_t$ is the next node to visit.

    \item \textit{Reward Function}: The agent receives a unit positive reward for every customer it visits. The agent is penalized if it does not complete the trip within the time horizon $U$. The reward function $R(s_t, a_t)$ for the state $s_t = \{s^n_t, s^v_t = (i_t, C_t, \tau_t)\}$ and action $a_t$ is given by:
    \begin{equation}
    R(s_t,a_t) = \begin{cases}
        +1, \ \text{if} \ i_t \in C_t \ \text{and} \ \tau_t > 0 \\
        -M, \ \text{if} \ i_t \neq \text{depot} \ \text{and} \ \tau_t \leq 0 \\
        0, \ \text{otherwise}
    \end{cases}
    \end{equation}
    where $U$ is the time horizon, $\tau_t$ is the remaining time and $M$ is the failure penalty that the agent receives if it fails to complete the trip within $U$. We heuristically compute the value of penalty $M$ as the total number of customers in the routing instance: $M = |\mathcal{C}|$.
\end{enumerate}

\subsection{Routing Network Embedding Module}
\label{net_emb_sec}

A reliable and effective vehicle routing method requires routing network representations that capture both node-level features and global routing context to enable optimal decision-making. However, existing deep RL methods lack context-aware global graph embeddings \cite{tang2023energy}.

We propose a novel routing network module using Graph Attention Networks (GAT) \cite{velivckovic2017graph} with Edge features and Cross-Attention mechanism. The GAT layers produce the node encoding vectors and we introduce a Cross-Attention-based global graph embedding vector to represent the global routing context.

\subsubsection{Graph Attention Networks with Edge Features (GAT-Edge).}

Consider a graph $G=(V,E)$ with the node features $X$, the binary adjacency matrix $A$ and edge features $e$. The definitions of $X$, $A$ and $e$ for the VRP-FTH are provided in the MDP. A Graph Attention Networks (GAT) layer takes $A$, $X$ and $e$ as inputs. First, the GAT layer computes attention scores $\hat{\alpha}_{ij}$ for each edge $(i,j) \in E$:
\begin{equation}
    \quad \hat{\alpha}_{ij} = \text{ReLU}(a^T[WX_i||WX_j])
\end{equation}
where $W$ is a learnable weight matrix to linearly transform the initial node features $X$ and $a$ is a learnable weight vector for the the graph attention layer.  

To incorporate the edge features, the attention scores ($\hat{\alpha}_{ij}$) are multiplied with the edge features ($e_{ij}$) to produce the edge-induced attention score ($\alpha^e_{ij}$):
\begin{equation}
\quad \alpha^e_{ij} = \text{softmax}(\hat{\alpha}_{ij}*e_{ij})
\label{eq_alpha1}
\end{equation}

The edge-induced attention scores $\alpha^e_{ij}$ are used as weight values to combine the neighbouring node features and compute the node embedding vectors $h^{(1)}$. The node embedding vector $h^{(1)}_i$ for node $i$ is computed as:
\begin{equation}
   \quad  h^{(1)}_i = \sigma (\sum_{k \in \mathcal{N}(i)} \alpha^e_{ik}W X_k)
\end{equation}
where $\sigma$ is the sigmoid function and $\mathcal{N}(i)$ is the set of neighbouring nodes of node $i$.

We apply two GAT-Edge layers and use residual connections to compute the final node embedding vectors. The second GAT-Edge layer produces $h^{(2)}$. The final node embedding vectors $h$ is computed by residually adding $h^{(1)}$ with $h^{(2)}$:
\begin{equation}
    \quad h = h^{(1)} + h^{(2)}
\end{equation}

\subsubsection{Global Graph Embedding using Cross-Attention Mechanism.}
\label{Attention_subsec}

While GAT-Edge layers capture local node features, global routing context is also highly important. We introduce a Cross-Attention layer that combines GAT-Edge node embeddings with the current vehicle state (including $U$) to generate a global graph embedding.

In the proposed cross-attention layer, the query vector $Q_t$ is generated from the vehicle state ($s^v_t$) as shown in Equation \ref{query_eq} and the node embedding vectors $h$ act as both the key vectors and value vectors. The attention scores $\alpha^t$ are generated from the query vector and the node embedding vector as shown in Equation \ref{score_eq}. Finally, to compute the global network embedding vector $h^G$, we linearly combine the value vectors (i.e. node embedding vectors $h$) using the attention scores, which is shown in Equation \ref{lc_eq}:
\begin{equation}
    \quad Q_t = W_q \cdot s^v_t
    \label{query_eq}
\end{equation}
\begin{equation}
    \quad \alpha^t = \text{softmax}(Q_t \cdot h^T)
    \label{score_eq}
\end{equation}
\begin{equation}
    \quad h^G = \sum_k \alpha^t_k h_k
    \label{lc_eq}
\end{equation}
where $W_q$ is a learnable weight matrix and $h_k$ is the node embedding vector for node $k$. $h^G$ is the context-aware global graph embedding vector. The routing network embedding module architecture is shown in Figure \ref{PG_GAT_Edge_fig}.

Additionally, we compute an active customer embedding vector ($h^{C_t}$) to represent the set of active customer requests in the following way:
\begin{equation}
    \quad h^{C_t} = \sum_{j\in C_t} h_j
    \label{customer_emb_eq}
\end{equation}
where $C_t$ is the set of active customer requests.

\subsection{Compatibility and Masking Layer} 

We create a context vector $H^C$ by concatenating the graph embedding vector $h^G$, the current node embedding vector $h_{i_t}$, the active customer requests embedding vector $h^{C_t}$ and the linear embedding the remaining time horizon $\tau$: $H^C = [h^G, h_{i_t}, h^{C_t}, \tau]$.

We compute the compatibility of the context vector $H^C$ and all the node in the routing network through a single-head attention layer:
\begin{equation}
    \quad u_i = C \cdot \text{tanh}(\frac{q^T k_i}{\sqrt{d_k}})
    \label{compat_eq}
\end{equation}
where $q = W^Q H^C$, $k_i = W^K h_i$, $W^Q$ and $W^K$ are learnable parameters, $d_k$ is the length of the query vector $q$ and $C$ is a constant.

To get the probabilities for each node from the compatibility values, we use a mask and a softmax activation layer \cite{tang2023energy}. At each timestep, the masking rules are: 
\begin{enumerate}
    \item All the previously visited nodes are masked.
    \item An unvisited node $v$ is masked if the agent cannot visit $v$ and return directly to depot within the remaining time horizon. 
\end{enumerate}

The probability vector for visiting different nodes are calculated in the following way: 
\begin{equation}
    \quad p_i = \text{softmax}(u_i+Z \cdot Mask_i)
    \label{mask_softmax_eq}
\end{equation}
where $p_i$ is the probability to visit node $i$, $Mask_i$ is the masking vector for node $i$ and $Z$ is a large negative number.

\subsection{Policy Gradient-based Routing Method}

The routing agent is implemented using a policy network that incorporates the routing network embedding module. The policy network includes a GAT-Edge and Cross-Attention-based embedding module, a context vector, and a compatibility layer. The network state $s^n_t$ and vehicle state $s^v_t$ serve as inputs. The embedding module produces node embeddings ($h$) and a global graph embedding ($h^G$), which are combined with active customer and remaining time embeddings to form the context vector $H^C$. The compatibility layer then maps $H^C$ to the policy $\pi(s_t, a_t)$.

We train the policy network using the REINFORCE Policy Gradient (PG) algorithm \cite{williams1992simple}. The policy gradient loss function is: 
\begin{equation}
\quad J(\theta) = \mathds{E}_\pi [G_t \ \text{log}(\pi(s_t|a_t;\theta))]
\end{equation}
where $\theta$ is the set of policy network weights and $G_t$ is the return or reward-to-go from the state-action pair $(s_t, a_t)$. The gradient of the loss is computed as:
\begin{equation}
\quad \nabla_{\theta} J(\theta) = \mathds{E}_\pi [G_t \ \nabla \text{log}(\pi(s_t|a_t;\theta))]
\end{equation}

Figure \ref{PG_GAT_Edge_fig} depicts our proposed policy network-based routing method architecture. We refer to our proposed routing method as Policy Gradient with GAT-Edge (PG-GAT-Edge).

\section{Experimental Results}


\begin{table*}[t]
\centering
\scalebox{0.8}{
\begin{tabular}{|l|cccc|}
\hline
\multicolumn{1}{|c|}{\multirow{3}{*}{\textbf{Method}}}                                         & \multicolumn{4}{c|}{\textbf{Customers Served (\%)}}                                                                                 \\ \cline{2-5} 
\multicolumn{1}{|c|}{}                                                                         & \multicolumn{2}{c|}{\textbf{Determinstic}}                                 & \multicolumn{2}{c|}{\textbf{Stochastic}}               \\ \cline{2-5} 
\multicolumn{1}{|c|}{}                                                                         & \multicolumn{1}{c|}{\textbf{EN-50}} & \multicolumn{1}{c|}{\textbf{EN-100}} & \multicolumn{1}{c|}{\textbf{EN-50}} & \textbf{EN-100} \\ \hline
GA \cite{maroof_ga_2024}                                                   & \multicolumn{1}{c|}{48.8}           & \multicolumn{1}{c|}{25.8}            & \multicolumn{1}{c|}{--}             & --              \\ \hline
VNS \cite{urrutia2021variable}                                               & \multicolumn{1}{c|}{37.8}           & \multicolumn{1}{c|}{19.9}            & \multicolumn{1}{c|}{--}             & --              \\
\hhline{|=|=|=|=|=|} DRL-Transformer \cite{tang2023energy}  & \multicolumn{1}{c|}{56.8}           & \multicolumn{1}{c|}{24.8}            & \multicolumn{1}{c|}{52.4}           & 26.2            \\ \hline
DRL-S2V \cite{lin2021deep}                                                   & \multicolumn{1}{c|}{41.1}           & \multicolumn{1}{c|}{17.5}            & \multicolumn{1}{c|}{23.7}           & 18.4            \\
\hhline{|=|=|=|=|=|} \textbf{PG-GAT-Edge (Proposed Method)} & \multicolumn{1}{c|}{\textbf{75.2}}  & \multicolumn{1}{c|}{\textbf{48.0}}   & \multicolumn{1}{c|}{\textbf{71.3}}  & \textbf{40.9}   \\ \hline
\end{tabular}}
\caption{Customer Service Rate (\%) for Different Routing Methods on Euclidean Routing Networks}
\label{Euclidean_results_table}
\end{table*}

\begin{table*}[t]
\centering
\scalebox{0.8}{
\begin{tabular}{|l|cccccc|}
\hline
\multicolumn{1}{|c|}{\multirow{3}{*}{\textbf{Method}}}                                             & \multicolumn{6}{c|}{\textbf{Customers Served (\%)}}                                                                                                                                                                     \\ \cline{2-7} 
\multicolumn{1}{|c|}{}                                                                             & \multicolumn{3}{c|}{\textbf{Deterministic}}                                                                          & \multicolumn{3}{c|}{\textbf{Stochastic}}                                                         \\ \cline{2-7} 
\multicolumn{1}{|c|}{}                                                                             & \multicolumn{1}{c|}{\textbf{EMA-50}} & \multicolumn{1}{c|}{\textbf{EMA-100}} & \multicolumn{1}{c|}{\textbf{EMA-150}} & \multicolumn{1}{c|}{\textbf{EMA-50}} & \multicolumn{1}{c|}{\textbf{EMA-100}} & \textbf{EMA-150} \\ \hline
GA \cite{maroof_ga_2024}                                                       & \multicolumn{1}{c|}{72.9}            & \multicolumn{1}{c|}{39.1}             & \multicolumn{1}{c|}{26.9}             & \multicolumn{1}{c|}{--}              & \multicolumn{1}{c|}{--}               & --               \\ \hline
VNS \cite{urrutia2021variable}                                                   & \multicolumn{1}{c|}{61.1}            & \multicolumn{1}{c|}{32.5}             & \multicolumn{1}{c|}{21.8}             & \multicolumn{1}{c|}{--}              & \multicolumn{1}{c|}{--}               & --               \\
\hhline{|=|=|=|=|=|=|=|} DRL-Transformer \cite{tang2023energy}  & \multicolumn{1}{c|}{74.1}            & \multicolumn{1}{c|}{46.1}             & \multicolumn{1}{c|}{27.3}             & \multicolumn{1}{c|}{28.7}            & \multicolumn{1}{c|}{22.0}             & 33.6             \\ \hline
DRL-S2V \cite{lin2021deep}                                                       & \multicolumn{1}{c|}{60.7}            & \multicolumn{1}{c|}{37.4}             & \multicolumn{1}{c|}{21.7}             & \multicolumn{1}{c|}{48.0}            & \multicolumn{1}{c|}{22.8}             & 31.6             \\ \cline{2-7} 
\hhline{|=|=|=|=|=|=|=|} \textbf{PG-GAT-Edge (Proposed Method)} & \multicolumn{1}{c|}{\textbf{86.6}}   & \multicolumn{1}{c|}{\textbf{57.5}}    & \multicolumn{1}{c|}{\textbf{41.7}}    & \multicolumn{1}{c|}{\textbf{94.5}}   & \multicolumn{1}{c|}{\textbf{49.8}}    & \textbf{56.2}    \\ \hline
\end{tabular}}
\caption{Customer Service Rate (\%) for Different Routing Methods on EMA highway networks}
\label{EMA_results_table}
\end{table*}
\begin{table*}[htbp!]
\centering
\scalebox{0.8}{
\begin{tabular}{|l|cc|}
\hline
\multicolumn{1}{|c|}{\multirow{2}{*}{\textbf{Method}}}                                     & \multicolumn{2}{c|}{\textbf{Customers Served (\%)}}             \\ \cline{2-3} 
\multicolumn{1}{|c|}{}                                                                     & \multicolumn{1}{c|}{\textbf{Vienna-160}} & \textbf{Vienna-300} \\ \hline
MKA \cite{Zhang_Luo_Florio_Van_Woensel_2023}                        & \multicolumn{1}{c|}{27.7}                & 13.9                \\
\hhline{|=|=|=|} DRL-Transformer \cite{tang2023energy}  & \multicolumn{1}{c|}{42.3}                & 25.8                \\ \hline
DRL-S2V \cite{lin2021deep}                                               & \multicolumn{1}{c|}{27.4}                & 12.5                \\
\hhline{|=|=|=|} \textbf{PG-GAT-Edge (Proposed Method)} & \multicolumn{1}{c|}{\textbf{47.9}}       & \textbf{28.2}       \\ \hline
\end{tabular}}
\caption{Customer Service Rate (\%) for Different Routing Methods on the Stochastic Vienna City Routing Networks}
\label{Vienna_results_table}
\end{table*}

\begin{table*}[htbp!]
\centering
\scalebox{0.8}{
\begin{tabular}{|l|ccccc|}
\hline
\multicolumn{1}{|c|}{\multirow{2}{*}{\textbf{Method}}}              & \multicolumn{5}{c|}{\textbf{Solution Time (s)}}                                                                                                                                        \\ \cline{2-6} 
\multicolumn{1}{|c|}{}                                              & \multicolumn{1}{c|}{\textbf{EMA-50}} & \multicolumn{1}{c|}{\textbf{EMA-100}} & \multicolumn{1}{c|}{\textbf{EMA-150}} & \multicolumn{1}{c|}{\textbf{Vienna-160}} & \textbf{Vienna-300} \\ \hline
MKA \cite{Zhang_Luo_Florio_Van_Woensel_2023} & \multicolumn{1}{c|}{--}              & \multicolumn{1}{c|}{--}               & \multicolumn{1}{c|}{--}               & \multicolumn{1}{c|}{26.59}               & 70.1                \\ \hline
GA \cite{maroof_ga_2024}                        & \multicolumn{1}{c|}{6.79}            & \multicolumn{1}{c|}{7.52}             & \multicolumn{1}{c|}{7.88}             & \multicolumn{1}{c|}{--}                  & --                  \\ \hline
VNS \cite{urrutia2021variable}                    & \multicolumn{1}{c|}{1.13}            & \multicolumn{1}{c|}{1.34}             & \multicolumn{1}{c|}{2.4}              & \multicolumn{1}{c|}{--}                  & --                  \\ \hline
DRL-Transformer \cite{tang2023energy}             & \multicolumn{1}{c|}{0.58}            & \multicolumn{1}{c|}{0.77}             & \multicolumn{1}{c|}{1.88}             & \multicolumn{1}{c|}{1.40}                & 3.48                \\ \hline
DRL-S2V \cite{lin2021deep}                        & \multicolumn{1}{c|}{0.46}            & \multicolumn{1}{c|}{0.63}             & \multicolumn{1}{c|}{1.77}             & \multicolumn{1}{c|}{0.92}                & 2.94                \\ \hline
\textbf{PG-GAT-Edge}                                               & \multicolumn{1}{c|}{\textbf{0.45}}   & \multicolumn{1}{c|}{\textbf{0.63}}    & \multicolumn{1}{c|}{\textbf{1.03}}    & \multicolumn{1}{c|}{\textbf{0.78}}       & \textbf{1.85}       \\ \hline
\end{tabular}}
\caption{Average Solution Time on Different Routing Networks}
\label{solution_time_table}
\end{table*}

In this section, we present the routing network instances, the experimental setup and the results of our experiments.

\subsection{Routing Network Datasets}

We have used synthetic Euclidean networks \cite{lin2021deep}, Vienna city networks \cite{Zhang_Luo_Florio_Van_Woensel_2023}, and newly created Eastern Massachusetts highway networks for our experiments. Our routing network datasets are as follows: 
\begin{enumerate}
    \item \textit{Euclidean Routing Network (EN)}: We generated random Euclidean routing networks comprising 50 and 100 customers (EN-50 and EN-100) following the methodology of \citet{lin2021deep}. Two network types were considered: deterministic (all customer requests are known at the start), and stochastic (half of the requests arrive during the trip). The time horizons were set to 24 hours for deterministic and 18 hours for stochastic instances.
    \item \textit{Eastern Massachusetts Highway Road Network (EMA Network)}: We created the Eastern Massachusetts Highway routing instances from the OpenStreetMap. The EMA network dataset contains 50-customer instances (EMA-50), 100-customer instances (EMA-100) and 150-customer instances (EMA-150). We created both deterministic (all customers are known at the start) and stochastic instances (half of the requests arrive during the trip). The time horizon ($U$) is 24 hours.
    \item \textit{Vienna City Network}: Following the experimental setup provided by \citet{Zhang_Luo_Florio_Van_Woensel_2023}, we generated the Vienna city routing instances from the Vienna city network, with 160 customers (Vienna-160) and 300 customers (Vienna-300). The time horizon ($U$) is 10 hours.
\end{enumerate} 

\subsection{Baseline Models}

We have compared our proposed routing method with five existing routing methods:
\begin{enumerate}
    \item \textit{Genetic Algorithm (GA)}: We implemented the Genetic Algorithm to solve the VRP-FTH as presented in \citet{maroof_ga_2024}. We considered the following hyper-parameters: population size = 100, number of generations = 1000, mutation rate = 0.01. 
    \item \textit{Variable Neighbourhood Search (VNS)}: We implemented the Variable Neighbourhood Search algorithm as proposed in \citet{urrutia2021variable} to solve the vehicle routing problem with finite horizon. 
    \item \textit{Mulitple Knapsack-based Approximation (MKA)}: \citet{Zhang_Luo_Florio_Van_Woensel_2023} proposed a multiple knapsack approximation-based approach for the VRP-FTH, combining a potential-based offline planner with an online knapsack method. We compare our method with this method only on the Vienna city network instances.
    \item \textit{Deep RL with Transformer (DRL-Transformer)}: The deep RL with transformer method was proposed by \citet{tang2023energy}. This method uses transformer to encode the routing network information. They use the policy gradient algorithm to train the agent.
    \item \textit{Deep RL with Structure2Vec (DRL-S2V)}: \citet{lin2021deep} proposed the DRL-S2V method which includes a structure2vec model to compute the node embedding vectors. They have also used the policy gradient algorithm.
\end{enumerate}

\subsection{Experimental Setting}

In our policy network, the network embedding module contains two GAT-Edge layers with 4 attention heads and 64-dimensional embedding vectors and the context vector ($H^C$) is projected into a 256-dimensional vector before computing the compatibility layer.

\subsection{Results}

\begin{figure*}[t]
\centering
    \begin{subfigure}{0.40\textwidth}
    \centering
        \includegraphics[width=\linewidth]{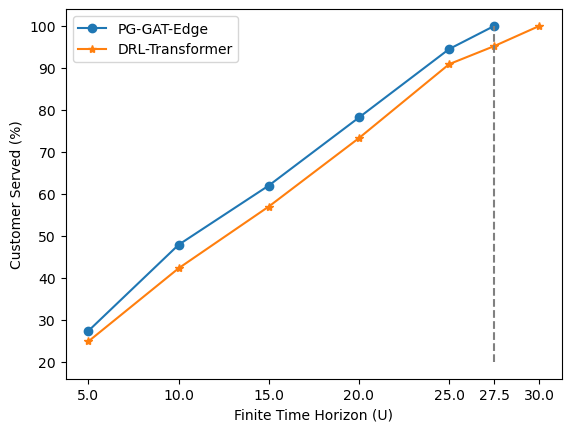}
        \caption{Vienna-160}
        \label{varying_horizon_Vienna_fig}
    \end{subfigure}%
    \hspace{1cm}
    \begin{subfigure}{0.40\textwidth}
    \centering
        \includegraphics[width=\linewidth]{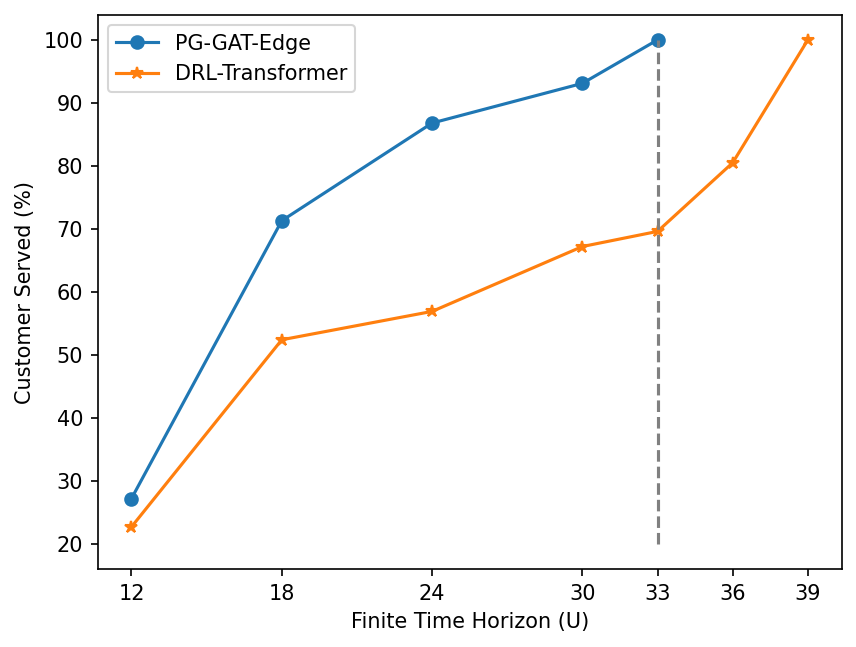}
        \caption{Euclidean Network-50}
        \label{varying_horizon_EN_fig}
    \end{subfigure}
    \caption{Customer Service Rate (\%) under different time horizon ($U$, in hours) for PG-GAT-Edge (Our method) and DRL-Transformer \cite{tang2023energy} on Vienna-160 and EN-50}
    \label{varying_horizon_fig}
\end{figure*}

Table \ref{Euclidean_results_table} presents the customer service rate results of the PG-GAT-Edge method and other baseline methods on the Euclidean routing instances. The PG-GAT-Edge method improves the customer service by 59.2\% on the deterministic Euclidean instances and by 46.0\% on the stochastic instances compared to the other methods.

Table \ref{EMA_results_table} shows the customer service rate results of different routing methods on the EMA highway routing instances. The PG-GAT-Edge method outperforms all the other methods on each of the instances. We present the customer service results on the Vienna city routing instances in Table \ref{Vienna_results_table}. On the Vienna stochastic instances, we observe that our proposed method achieves 72.9\% more customer service rate on Vienna-160 instances and 102.8\% more customer service rate on the Vienna-300 instances than the Multiple Knapsack Approximation method \cite{Zhang_Luo_Florio_Van_Woensel_2023}.  

Table \ref{solution_time_table} presents the solution time (i.e. decision time) of different method on the EMA instances and the Vienna instances. It is observed that the solution time of PG-GAT-Edge is significantly less than most of the other methods. Additionally, as the network size grows, the solution time gap between our method and the other methods increase.

Our experimental results demonstrate that the proposed method improves customer service rates on both real-world and randomly generated Euclidean networks. These results highlight the importance of the graph embedding module, as most of the existing deep RL methods using only node coordinates fail to properly capture the network structure. Our GAT-Edge-based routing network embedding module effectively understands the underlying graph structure and the edge features. Additionally, incorporating the edge features significantly improves our model's performance.

\subsection{Effects of Different Finite Horizons}

We compared the customer service rates under varying time horizons ($U$) for DRL-Transformer and PG-GAT-Edge on Vienna-160 and EN-50 instances. As shown in Figure \ref{varying_horizon_fig}, PG-GAT-Edge consistently outperforms DRL-Transformer across all $U$ values. Figure \ref{varying_horizon_EN_fig} shows this performance gap widens with increasing $U$.

The smallest value of the finite horizon ($U$) at which the routing agent is able to serve 100\% of the customer requests, represents the minimum time required to serve all the customers and complete the trip. Therefore, by suitably modifying the MDP in our VRP framework, the proposed routing method can be extended to other variants of the Vehicle Routing Problem, such as time-minimization VRP, distance-minimization VRP and VRP with time windows.

\subsection{Ablation Study}

\begin{table}[htbp]
\centering
\scalebox{0.8}{
\begin{tabular}{|l|c|}
\hline
\multicolumn{1}{|c|}{\textbf{Method}}                                                              & \begin{tabular}[c]{@{}l@{}}\textbf{Customers}\\ \textbf{Served (\%)}\end{tabular} \\ \hline
PG-GAT without Edge Features                                                              & 83.2                  \\ \hline
PG-GAT-Edge without Global Graph Embedding                                                             & 76.4                  \\ \hline
\begin{tabular}[c]{@{}l@{}}PG-GAT-Edge without Finite Horizon in Embedding\end{tabular} & 78.2                  \\ \hline
\textbf{PG-GAT-Edge (Proposed Method)}                                                             & \textbf{86.6}         \\ \hline
\end{tabular}}
\caption{Ablation Study for Edge Features, Global Graph Embedding and Finite Horizon in Graph Embedding on the EMA-50 instances}
\label{ablation_study}
\end{table}

Table \ref{ablation_study} presents an ablation study for the edge features, the global graph embedding module and the incorporation of finite horizon in the embedding module. We observe that the incorporation of edge features in the GAT layers improved the routing results by 4.1\%.

The customer service rate of PG-GAT-Edge without the global graph embedding (GE) module is 11.8\% lower than the proposed method. Similarly, we observe that the incorporation of the finite horizon in the graph embedding module improved the customer service rate by 10.7\%.

These ablation study results prove that the proposed Cross-Attention-based global graph embedding module and the inclusion of finite horizon in the embedding help our proposed model achieve better customer service rate.

\section{Conclusion}

In this paper, we proposed a deep RL-based routing method with a GAT-Edge-based network representation module to solve the VRP-FTH. We created a novel routing network embedding module using GAT-Edge and Cross-Attention mechanism, which is capable of extracting important local network features as wells as a global network representation. The incorporation of the finite horizon (and other current vehicle state components) in the graph embedding module enabled the agent take better routing decisions. We integrated the graph embedding module with a policy network-based routing method. The main results of this paper can be summerised as:
\begin{enumerate}
    \item Our method improves the customer service rate by 57.7\% on EMA networks and by 11.2\% on Vienna networks compared to the other methods.
    \item PG-GAT-Edge improves customer service by 52.6\% on Euclidean instances compared to other methods.
    \item The solution time of our proposed method is 90.64\% less than genetic algorithm and 35.4\% less than the DRL-Transformer method.
\end{enumerate}

In future work, research efforts could be made to extend the proposed deep RL-based framework for vehicle fleet routing.

\bibliography{references}
\end{document}